%% file: main.tex
\documentclass[lettersize,journal]{IEEEtran}
\usepackage{amsmath,amsfonts}
\usepackage{algorithmic}
\usepackage{algorithm}
\usepackage{array}
\usepackage[caption=false,font=normalsize,labelfont=sf,textfont=sf]{subfig}
\usepackage{textcomp}
\usepackage{stfloats}
\usepackage{url}
\usepackage{verbatim}
\usepackage{graphicx}
\usepackage{cite}
\usepackage{multirow}
\usepackage{colortbl}
\usepackage[dvipsnames]{xcolor}
\usepackage{pdfpages}
\usepackage[colorlinks,linkcolor=blue]{hyperref}
\begin{document}

\title{BViT: Broad Attention based Vision Transformer}

\author{Nannan Li,~\IEEEmembership{Graduate Student Member,~IEEE}, Yaran Chen,~\IEEEmembership{Member,~IEEE}, Weifan Li,~\IEEEmembership{Graduate Student Member,~IEEE}, Zixiang Ding,~\IEEEmembership{Member,~IEEE}, Dongbin Zhao,~\IEEEmembership{Fellow,~IEEE}, Shuai Nie
        % <-this % stops a space
\thanks{This work was supported in part by the National Natural Science Foundation of China (NSFC) under Grant 62173324 and 62006223, and the International Partnership Program of the Chinese Academy of Sciences under Grant 104GJHZ2022013GC.}% <-this % stops a space
\thanks{N. Li, Y. Chen, W. Li, Z. Ding and D. Zhao are with the State Key Laboratory of Multimodal Artificial Intelligence Systems, Institute of Automation, Chinese Academy of Sciences, Beijing 100190, China and also with the School of Artificial Intelligence, University of Chinese Academy of Sciences, Beijing 100049, China. S. Nie is with the National Laboratory of Pattern Recognition, Institute of Automation, Chinese Academy of Sciences, Beijing 100190, China (email : linannan2017@ia.ac.cn, chenyaran2013@ia.ac.cn, liweifan2018@ia.ac.cn, dingzixiang2018@ia.ac.cn, dongbin.zhao@ia.ac.cn, shuai.nie@nlpr.ia.ac.cn).}}

\markboth{IEEE Transactions on Neural Networks and Learning Systems,~Vol.~, No.~, ~2022}%
{Shell \MakeLowercase{\textit{et al.}}: A Sample Article Using IEEEtran.cls for IEEE Journals}


\maketitle

\input{abstract}
\input{introduction}

\input{related}

\input{method}

\input{experiment}
\input{conclusion}

\bibliographystyle{IEEEtran}
\bibliography{IEEEexample}

\begin{IEEEbiography}[{\includegraphics[width=1in,height=1.4in]{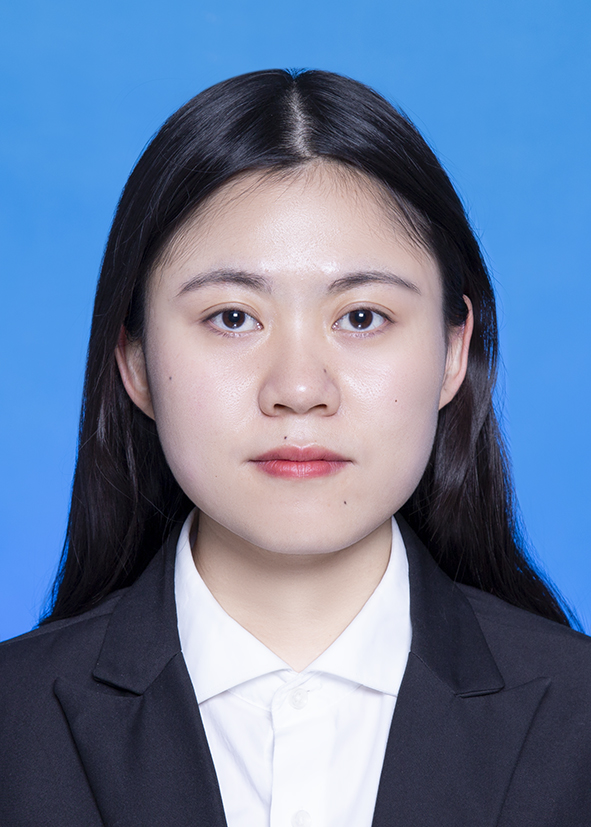}}]{Nannan Li}
received the B.S. degree in the School of Automation Engineering, University of Electronic Science and Technology of China, Sichuan, China, 2017. She is currently pursuing a Ph.D. degree in control theory and control engineering at The State Key Laboratory of Multimodal Artificial Intelligence Systems, Institute of Automation, Chinese Academy of Sciences, Beijing, China. Her research interests include computer vision and neural architecture search.
\end{IEEEbiography}
\begin{IEEEbiography}[{\includegraphics[width=1in,height=1.4in]{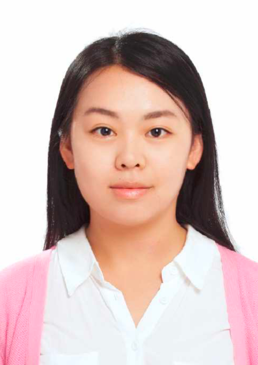}}]{Yaran Chen}
Associate Professor, Institute of Automation, Chinese Academy of Sciences, China. She received the Ph.D. degree from Institute of Automation, Chinese Academy of Sciences in 2018. She is currently an associate professor at The State Key Laboratory of Multimodal Artificial Intelligence Systems, Institute of Automation, Chinese Academy of Sciences, Beijing, and also with the College of Artificial Intelligence, University of Chinese Academy of Sciences, China. Her research interests include deep learning, neural architecture search, deep reinforcement learning and autonomous driving.
\end{IEEEbiography}
\begin{IEEEbiography}[{\includegraphics[width=1in,height=1.4in]{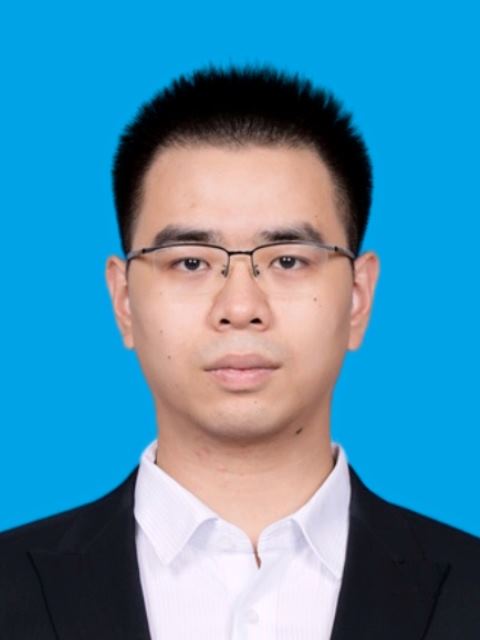}}]{Weifan Li} received the B.S. degree in materials science and engineering from Chongqing University, Chongqing, China, in 2015, and the M.S. degree in automation from Fuzhou University, Fuzhou, Fujian, China, in 2018. He is currently pursuing the Ph.D. degree in control theory and control engineering with The State Key Laboratory of Multimodal Artificial Intelligence Systems, Institute of Automation, Chinese Academy of Sciences, Beijing, China. His current research interests include reinforcement learning, deep learning, and game AI. 
\end{IEEEbiography}
\begin{IEEEbiography}[{\includegraphics[width=1in,height=1.4in]{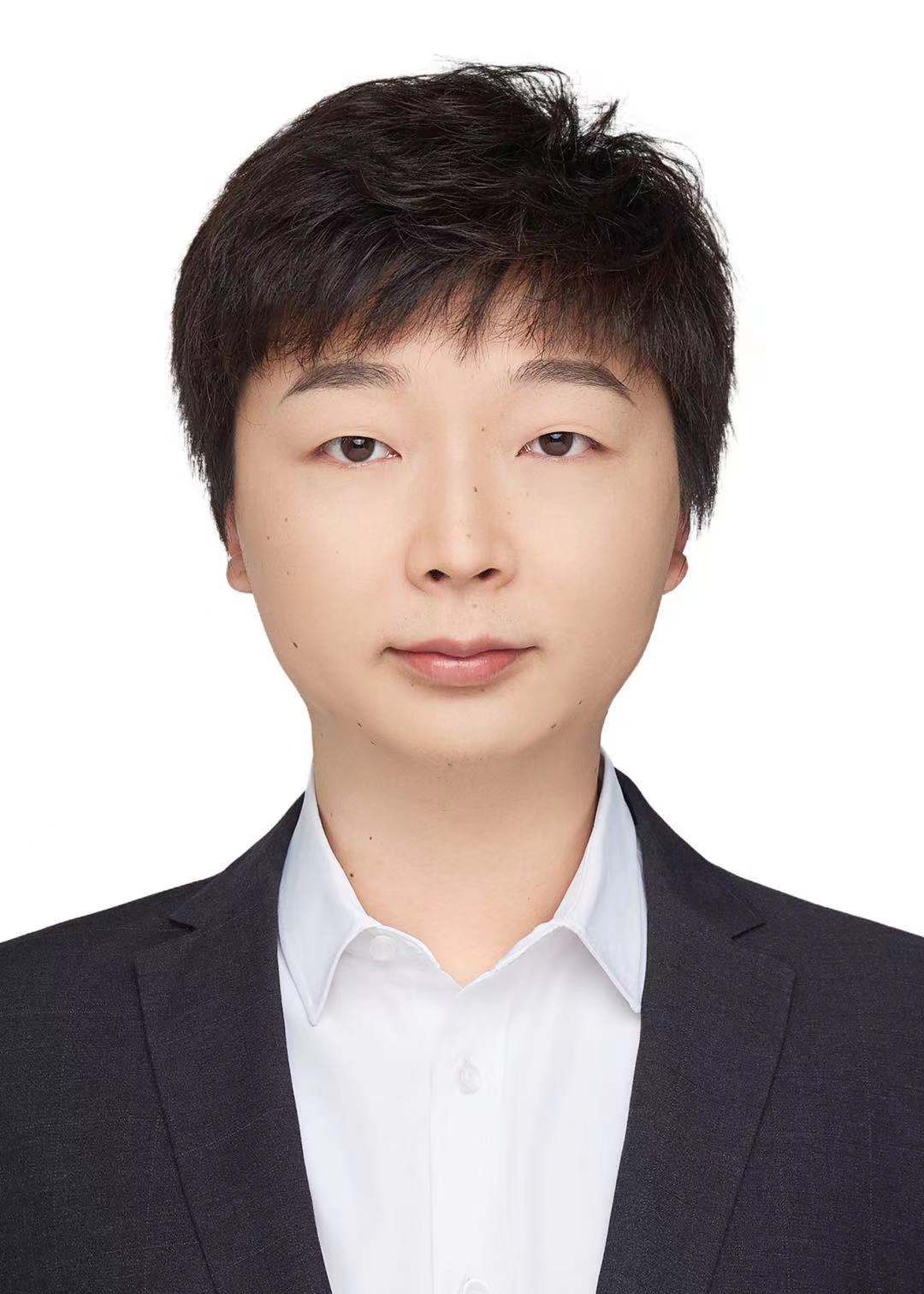}}]{Zixiang Ding}
received the M.E. degree in the School of Information and Electrical Engineering, Shandong Jianzhu University, Shandong, China, 2018. He is currently pursuing a Ph.D. degree in computer applications at The State Key Laboratory of Multimodal Artificial Intelligence Systems, Institute of Automation, Chinese Academy of Sciences, Beijing, China. His research interests include computer vision, neural architecture search and deep reinforcement learning.
\end{IEEEbiography}
\begin{IEEEbiography}[{\includegraphics[width=1in,height=1.4in]{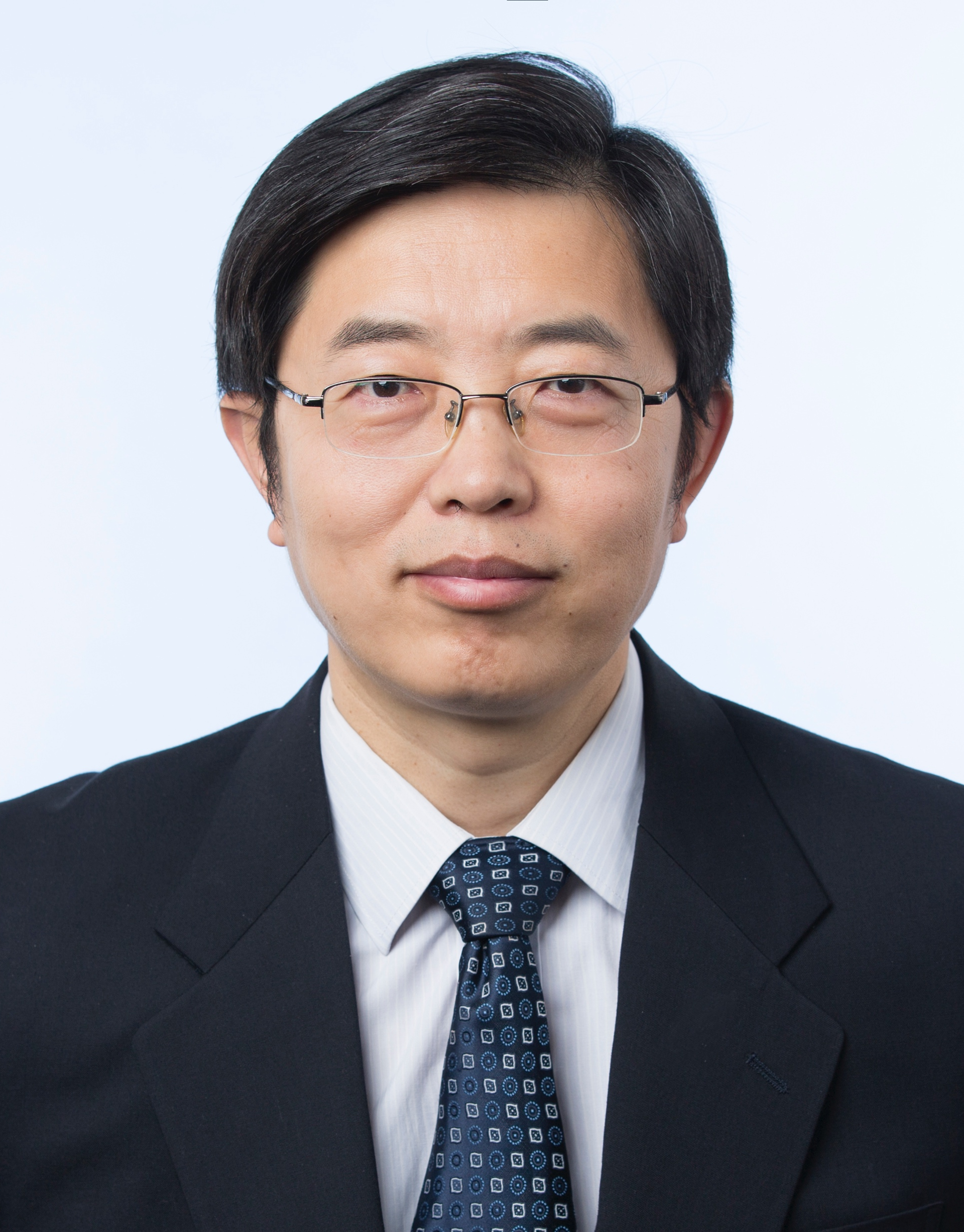}}]{Dongbin Zhao}
(M’06-SM’10-F’20) received the Ph.D. degrees from Harbin Institute of Technology, Harbin, China, in 2000. He is now a professor with Institute of Automation, Chinese Academy of Sciences, and also with the University of Chinese Academy of Sciences, China. He has published 6 books, and over 100 international journal papers. His current research interests are in the area of deep reinforcement learning, computational intelligence, autonomous driving, game artificial intelligence, robotics, etc.

Dr. Zhao serves as the Associate Editor of \textit{IEEE Transactions on Neural Networks and Learning Systems, IEEE Transactions on Cybernetics, IEEE Transactions on Artificial Intelligence}, etc. He is the chair of Distinguished Lecture Program, and was the Chair of \textit{Technical Activities Strategic Planning Sub-Committee (2019), Beijing Chapter (2017-2018), Adaptive Dynamic Programming and Reinforcement Learning Technical Committee (2015-2016)} of IEEE Computational Intelligence Society (CIS). He works as several guest editors of renowned international journals. He is involved in organizing many international conferences. He is an IEEE Fellow.
\end{IEEEbiography}
\begin{IEEEbiography}[{\includegraphics[width=1in,height=1.4in]{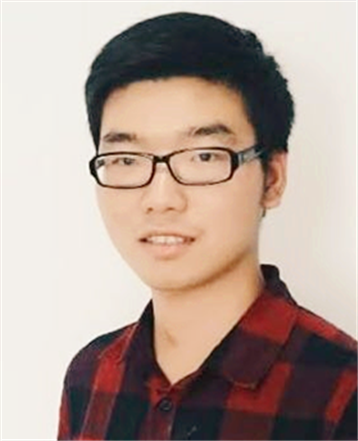}}]{Shuai Nie}
received the Ph.D degree in pattern recognition and intelligent systems from the National Key Laboratory of Pattern Recognition, Institute of Automation, Chinese Academy of Sciences, Beijing, China, in 2018.
He is an Associate Professor with the National Key Laboratory of Pattern Recognition, Institute of Automation, Chinese Academy of Sciences. His research interests include speech recognition, speech separation/enhancement, deep learning and large language models.
\end{IEEEbiography}

\end{document}

%% file: abstract.tex
\begin{abstract}
Recent works have demonstrated that transformer can achieve promising performance in computer vision, by exploiting the relationship among image patches with self-attention. While they only consider the attention in a single feature layer, but ignore the complementarity of attention in different layers. In this paper, we propose broad attention to improve the performance by incorporating the attention relationship of different layers for vision transformer, which is called BViT. The broad attention is implemented by broad connection and parameter-free attention. Broad connection of each transformer layer promotes the transmission and integration of information for BViT. Without introducing additional trainable parameters, parameter-free attention jointly 
focuses on the already available attention information in different layers for extracting useful information and building their relationship. 
Experiments on image classification tasks demonstrate that
BViT delivers superior accuracy of 75.0\%/81.6\% top-1 accuracy on ImageNet with 5M/22M parameters. Moreover, we transfer BViT to downstream object recognition benchmarks
to achieve 98.9\% and 89.9\% on CIFAR10 and CIFAR100 respectively that exceed ViT with fewer parameters. For the generalization test, the broad attention in Swin Transformer, T2T-ViT and LVT also brings an improvement of more than 1\%. To sum up, broad attention is promising to promote the performance of attention-based models.
Code and pre-trained models are available at \href{https://github.com/koala719/BViT}{https://github.com/DRL/BViT}.
\end{abstract}


%% file: introduction.tex
\section{Introduction}
\label{sec:intro}

\begin{figure}[t]
  \centering
   \includegraphics[width=1.0\linewidth]{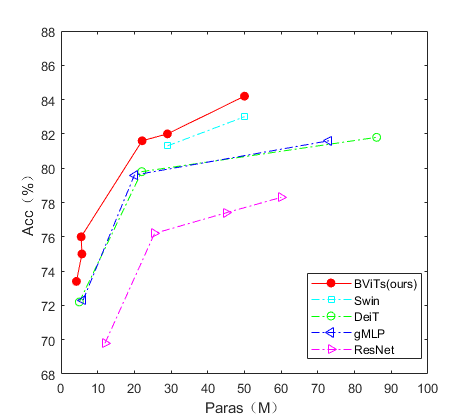}
           \caption{Parameters and accuracy on Imagenet of BViTs (i.e. BViT, BSwin, BLVT and BT2T-ViT) compared to transformer-based, MLP-based and  convolution-based models, such as DeiT~\cite{touvron2021training}, Swin~\cite{liu2021swin}, gMLP~\cite{liu2021pay} and ResNet~\cite{he2016deep}.}
   \label{fig:acc}
\end{figure}
\IEEEPARstart{T}{ransformer}~\cite{vaswani2017attention} has demonstrated impressive modeling capabilities and achieved state-of-the-art performance in natural language processing tasks. Recently, Vision Transformer (ViT)~\cite{DBLP:conf/iclr/DosovitskiyB0WZ21} has been proposed and made a breakthrough, which is widely believed to be expected to break the dominance of Convolutional Neural Network (CNN) in computer vision~\cite{he2016deep,lu2020cnn,mellouli2019morphological,liu2020stroke, li2022abcp, li2022heuristic}. ViT usually divides the whole image into many fixed-size patches containing local components and then exploits the relationship between them by self-attention, which can focus modeling capabilities on the information relevant to vision tasks, such as image classification~\cite{DBLP:conf/iclr/DosovitskiyB0WZ21, touvron2021training}, object detection~\cite{liu2021swin, carion2020end}, and semantic segmentation~\cite{strudel2021segmenter}.

Many works have demonstrated that the self-attention mechanism plays a critical role in the success of ViT. 
Even if they are far away, the relationship between local components in an image can be exploited by self-attention. Furthermore, multi-head attention can focus on useful local information and relationship in images from different views.
Many efforts have been devoted to designing self-attention to improve the capabilities of focusing on useful information and suppressing redundant information. 
Swin Transformer~\cite{liu2021swin} hierarchically performs self-attention on a shifted window of patches, which increases the connection between cross windows and has the flexibility of modeling on various scales. Tokens-to-Token ViT~\cite{yuan2021tokens} recursively assembles neighboring tokens into a token to extract the structure information of the image by self-attention incrementally. CvT~\cite{wu2021cvt} introduces convolution to vision transformer to achieve additional attention modeling of local spatial context. However, these efforts only consider the attention performed on feature maps from one layer but ignore that the combination of attention in different layers is helpful. 
Specifically, shallower layers focus on both local and global information, while deeper layers tend to focus on global information~\cite{DBLP:conf/iclr/DosovitskiyB0WZ21}. And the study on the difference between ViT and CNN~\cite{raghu2021vision} suggests that ViT performs poorly on limited datasets (e.g. ImageNet) due to inadequate attention to local features. Further, the Centered Kernel Alignment (CKA) similarity scores between shallower and deeper layers in ViT are higher than that in CNN~\cite{raghu2021vision}, which may exhibit that the architecture is redundant. In a word, similar layers can be pruned with minimal impact on performance~\cite{nguyen2020wide}. The combination of attention from different layers is promising to alleviate the above problems, not only allowing the deep layers to acquire local information, but also making fuller exploitation of features.
% To alleviate the above problems, we intend to design a new attention paradigm that combines attention in different layers.

It is widely accepted that nonlinear processing and pathway connections between different layers are the key to the success of deep neural networks. The multi-layer nonlinear operations provide the hierarchical feature extraction capability for the model. At the same time, the pathway connections between the different layers facilitate the transmission and integration of information, which can be used to combine the attention in different layers. Following we go back to deep CNN for consolidation of the above discussion.
ResNet~\cite{he2016deep} 
stacks more convolution layers to achieve a very deep network. Furthermore, it designs residual connections that effectively improve performance and develop the training process. DenseNet~\cite{huang2016deep} further increases the pathway connection, i.e. dense connection, which brings more improvement in performance. BNAS~\cite{ding2021bnas}, a neural architecture search method, also designs a broad connection, adding pathway connections among the shallower layers and the last layer for collecting features with different layers.
The revolutionary success of pathway connections on CNN proves its benefit to learning effective features. 
Therefore, it is reasonable to expect the combination and full exploitation of attention in different layers to be implemented by increasing the path connections. Concretely, combining attention in different layers facilitates i) the transmission and fusion of attention, which promotes attention to local information, and ii) the extraction of practical attention information, which mitigates model redundancy.

In this paper, we propose broad attention mechanism that can efficiently extract and utilize the knowledge in each transformer layer. In particular, we first integrate attention information in different transformer layers via a broad connection. Then we perform parameter-free attention on the integrated features mentioned above for extracting helpful contents and their structural relationship hierarchically.
It is to be noted that no additional trainable parameters are required since that parameter-free attention is executed on information already processed by self-attention in transformer layers. More significantly, broad attention is generic, providing attention-based models with the flexibility to introduce broad attention blocks for improved performance.

Based on broad attention, we present Broad attention based Vision Transformer, called BViT. BViT consists of two components: 1) BViT backbone, including multiple transformer layers, is to yield deep features; 2) broad attention block derives broad features by increasing the path connection of attention in different layers and extracting helpful information hierarchically without extra learnable parameters.
The integration of the BViT backbone and the broad attention block results in a model with superior image understanding capabilities.

The proposed BViT presents powerful performance on image classification benchmark tasks. The model resulting from our method exceeds transformer-based models ViT/DeiT~\cite{DBLP:conf/iclr/DosovitskiyB0WZ21, touvron2021training}, MLP-based models Mixer/gMLP~\cite{tolstikhin2021mlp,liu2021pay} and CNN-based models ResNet/RegNetY~\cite{he2016deep, radosavovic2020designing} with comparable parameters on ImageNet~\cite{deng2009imagenet} dataset. We also prove the transferability on different transfer benchmark datasets (CIFAR10/100~\cite{krizhevsky2009learning}) with pre-training on ImageNet. 
Further, to support that broad attention can be employed to attention-based models as a generic mechanism, we conduct experiments that apply broad attention to Swin Transformer~\cite{liu2021swin}, T2T-ViT~\cite{yuan2021tokens} and LVT~\cite{yang2022lite}. In particular, for both models, there is a positive 1\% rise in the result on ImageNet. The comparison of parameters and accuracy between our proposed methods and other typical models is shown in Fig.~\ref{fig:acc}. 

In summary, our contributions are outlined below:

\begin{itemize}
\item We propose a novel broad attention based vision transformer. Broad attention can extract effective features without any extra trainable parameters. The proposed BViT attains superior classification accuracy on ImageNet with about 2\% improvement compared to ViT and ResNet. Moreover, the pre-trained model can achieve comparable performance on downstream tasks, including CIFAR10 and CIFAR100.  
\item The visualization results demonstrate that broad attention makes the model attend more locally, which facilitates the performance on the limited datasets. And broad attention brings smaller feature similarity, which prevents model redundancy.
\item Broad attention mechanism performs well in generalizability, which can elevate the performance of attention-based models flexibly. The results of extensive experiments show that implementations on excellent and prevail attention-based models present favorable performance improvement.
\end{itemize}

%% file: related.tex
\section{Related Work}

\subsection{Vision Transformer}
Transformer~\cite{vaswani2017attention} was initially devised for natural language processing and has achieved leading performance. The key point of the transformer is to fetch the global dependencies of the input via a self-attention mechanism. Further, in an endeavor to explore more potential of transformers, recent research has been directed to the application of transformers in the field of computer vision, such as ViT~\cite{DBLP:conf/iclr/DosovitskiyB0WZ21}, DeiT~\cite{touvron2021training}. ViT~\cite{DBLP:conf/iclr/DosovitskiyB0WZ21} first showed the strength of pure transformer architecture on image classification with large-scale image datasets (i.e. ImageNet-21K~\cite{deng2009imagenet}, JFT-300M~\cite{sun2017revisiting}). ViT processes the image input into a sequence paradigm by the following steps: \romannumeral1) partitioning the image into fixed-size patches; \romannumeral2) applying linear embedding to fixed-size patches; \romannumeral3) adding the position embedding. With an extra learnable “classification token”, the sequence can be handled by a standard transformer encoder.

Based on the innovative ViT~\cite{DBLP:conf/iclr/DosovitskiyB0WZ21}, several studies are dedicated to improving the architecture design and enhancing the performance of ViT. DeiT~\cite{touvron2021training} performed data-efficient training and introduced a distillation token for more powerful knowledge distillation. Swin Transformer~\cite{liu2021swin} computed hierarchical feature representation via shifted window and attained linear computational complexity relative to the input resolution size. Tokens-to-Token ViT~\cite{yuan2021tokens} proposed a tokens-to-token process to aggregate surrounding tokens and enhance local information gradually. Autoformer~\cite{chen2021autoformer} searched high-performed transformer-based models through one-shot architecture search. The search space consists of embedding dimension, MLP ratio, depth, and so on. In addition, some works combined convolution and transformer architecture to exploit the advantages of both them~\cite{wu2021cvt, li2021bossnas, liu2021uninet}. The above-mentioned works have all achieved remarkable performance, but most of them neglected the exploitation of attention in different layers.
\subsection{Attention Mechanism}
The attention mechanism is first demonstrated to be helpful in computer vision. Google~\cite{mnih2014recurrent} performed attention mechanism on recurrent neural network for image classification. Subsequently, the attention mechanism is introduced into natural language processing~\cite{DBLP:journals/corr/BahdanauCB14,vaswani2017attention}. Transformer~\cite{vaswani2017attention} exhibited the power of the attention mechanism on natural language processing tasks and proposed the strong multi-head self-attention, which is widely used~\cite{zhao2020spatial,DBLP:conf/iclr/DosovitskiyB0WZ21,touvron2021training}.  Recently, attention mechanism has made a brilliant comeback in the image processing tasks, such as image classification~\cite{DBLP:conf/iclr/DosovitskiyB0WZ21,liu2021swin,touvron2021training,zhao2016deep}, object detection~\cite{wang2018non,cao2019gcnet}, semantic segmentation~\cite{fu2019dual,zhang2020feature}. Due to the robustness of the attention mechanism, there are many studies carried out on the attention mechanism.

The innovation of the attention mechanism is ongoing. Some works presented multi-level attention~\cite{yu2017multi,ma2019attnsense,li2017mam}. MLAN~\cite{yu2017multi} jointly leveraged visual attention and semantic attention to process visual question answering. MAM-RNN~\cite{li2017mam} includes frame-level attention layer and region-level attention layer, which can jointly focus on the notable regions in each frame and the
frames correlated to the target caption. Besides, HAN~\cite{yang2016hierarchical} made two levels
of attention mechanisms, i.e. word and sentence respectively, for differentially attending to content with different importance during the construction of the document representation. Unlike the above-mentioned works, we are interested in the information of attention that already exists in different layers of the model. A broader view of attention contributes to the acquisition of diverse features.

\subsection{Pathway Connection}
As the core component of deep learning, pathway connection facilitates the transmission of gradients and the availability of comprehensive features. AlexNet~\cite{krizhevsky2012imagenet} is regarded as the first truly deep convolutional neural network structure, which has made a breakthrough in large-scale image datasets. ResNet~\cite{he2016deep} 
added residual connections between convolutional layers, which effectively facilitates information transmission in very deep networks and alleviates the gradient dispersion problem of optimization. DenseNet~\cite{huang2016deep} further increased the pathway between layers by means of dense connections, which significantly strengthens the propagation and fusion of features and improves the modeling ability of deep networks.

As one of the connection types for pathway connection, broad connection can compensate for the lack of training efficiency and feature diversity in the deep network. Introducing broad connection, Ding et al. proposed Broad Neural Architecture Search (BNAS)~\cite{ding2021bnas} and its extended works~\cite{ding2022bnas}\cite{ding2021stacked}. BNAS~\cite{ding2021bnas} first presented and searched Broad Convolutional Neural Network (BCNN), and its search efficiency is leading in reinforcement learning based architecture search methods~\cite{pham2018efficient, zoph2018learning} and evolutionary algorithm based methods~\cite{chen2021modulenet,sun2019completely}. 
Stacked BNAS~\cite{ding2021stacked} developed the connection paradigm of BCNN, achieving performance improvement. The efficiency of BNAS confirms the advantages of the broad connection paradigm, efficient training and comprehensive features.

Broad connection helps extract rich information in different layers, while deep representations are more effective than shallow ones. Therefore, our method introduces broad connection without discarding deep features, which can obtain a wider variety of useful information with fewer sacrifices in model efficiency.

%% file: method.tex
\section{Methodology}
\begin{figure*}[t]
  \centering 
   \includegraphics[width=1.0\linewidth]{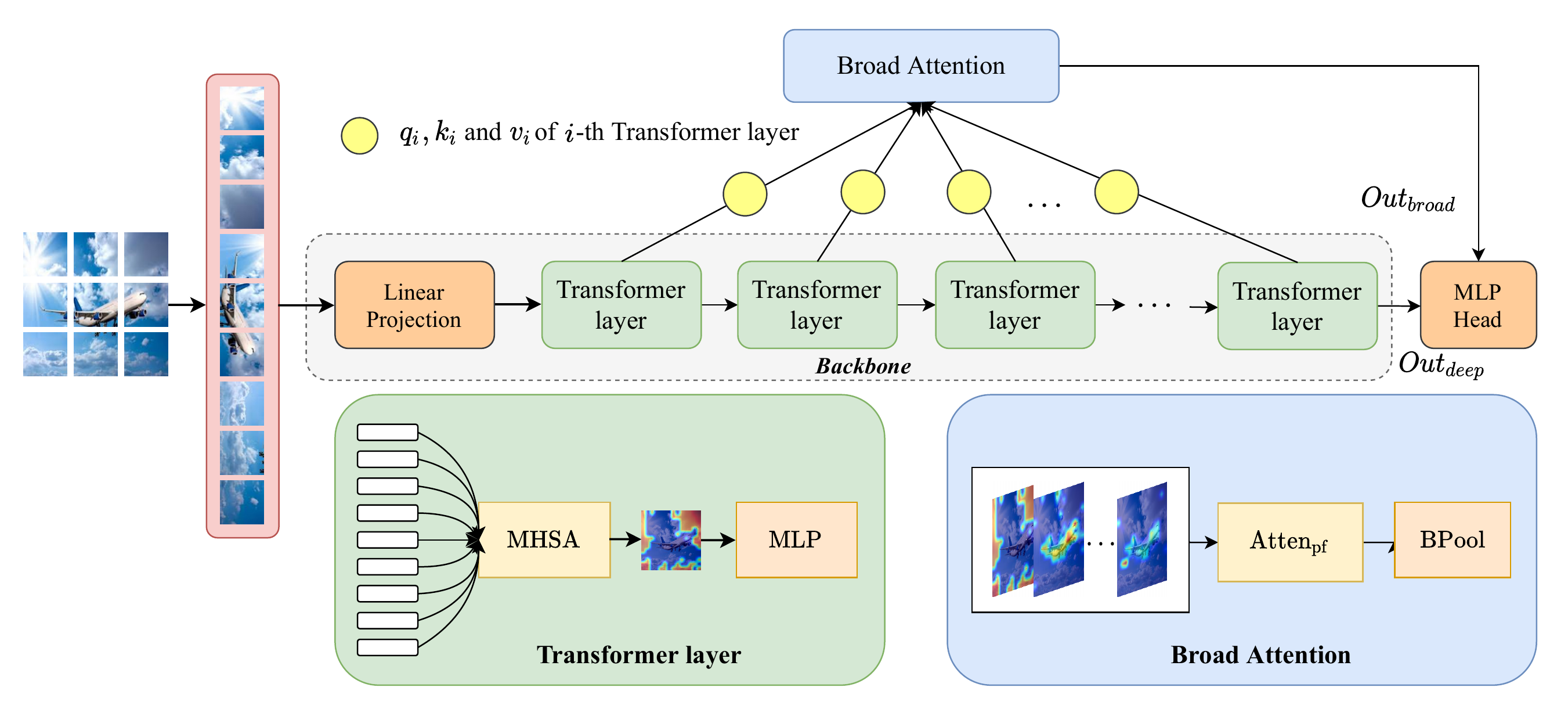}
   \caption{The overall architecture of BViT, which consists of BViT backbone (i.e. gray boxes) and broad attention (i.e. blue boxes). A patch feature input the backbone of BViT which consists several transformer layers, obtaining the feature $Out_{deep}$. The proposed broad attention mechanism extracts information from each transformer layer by broad connection, obtaining the diverse feature $Out_{broad}$.}
   \label{fig:overall}
\end{figure*}

As shown in Fig.~\ref{fig:overall}, the proposed model BViT mainly consists of two components: BViT backbone including multiple transformer layers (the gray boxes in Fig.~\ref{fig:overall}) for deep feature and broad attention (the blue box in Fig.~\ref{fig:overall}) for broad feature. BViT backbone obtains deep feature $Out_{deep}$ by calculating attention among patch features layer by layer. Then the attention in all transformer layers is collected by the proposed broad attention component with broad connection. Via parameter-free attention, we cannot only take full advantage of features in different layers jointly but also pay global attention on attention from each layer. Then the broad feature $Out_{broad}$ is obtained.

In the following section, we introduce the BViT backbone and the broad attention in detail.

\subsection{BViT Backbone}

Given a RGB image input $\mathbf{I} \in \mathbb{R}^{ H \times W \times C}$, we split it into non-overlapping and fixed-length patches $\mathbf{I_p} \in \mathbb{R}^{ N \times (P^2 \times C)}$. 
$P$ is the size of image patch, and $N = HW/P^2$ is the number of image patches. Our patch size $P$ is set to 16. Then image patches can be handled as a sequence, which can be processed directly by the standard transformer. To obtain the input $x_1 \in \mathbb{R}^{ N \times D}$ of the first transformer layer, a linear projection of flattened patches is employed for satisfying the consistent dimension $D$ across all layers of the transformer. Similar to ViT~\cite{DBLP:conf/iclr/DosovitskiyB0WZ21}, we also introduce trainable position embedding for retention of location information and classification token for image representation. With the classification token, the input is represented as $x_1 \in \mathbb{R}^{ (N+1) \times D}$.

After the necessary process of the image input, the data flow of the BViT architecture is divided into two orientations, i.e. BViT backbone and broad attention. As shown in Fig.~\ref{fig:overall}, BViT backbone (gray boxes) outputs the deep feature $Out_{deep}$ straightforwardly via several transformer layers.
A transformer layer includes two blocks: Multi-Head Self-Attention (MHSA) and Multi-Layer Perceptron (MLP). Besides, both MHSA and MLP blocks employ residual connections and apply LayerNorm (LN)~\cite{ba2016layer} before each block. Next, we detail the computation processes of MHSA and MLP.
% , especially MHSA which is quite helpful for understanding broad attention.

\textbf{Multi-Head Self-Attention}: MHSA attends to information at different positions using multi heads. Given the input tensor $x_i \in \mathbb{R}^{ (N+1) \times D}$ for $i$-th layer, 
the output of MHSA block is shown as 
\begin{equation}
\begin{aligned}
  \hat{z_i} = x_i + \text{MHSA}(x_i).
  \label{eq:h_z}
\end{aligned}
\end{equation}

\textbf{Multi-Layer Perceptron}: There are two fully connected layers in MLP block. And an activation function (e,g. GELU~\cite{hendrycks2016gaussian}) is between them. 
The output $z_i$ in $i$-th layer is formulated as
\begin{equation}
\begin{aligned}
  z_i = \hat{z_i} + \text{MLP}(\hat{z_i}).
  \label{eq:h_z}
\end{aligned}
\end{equation}
The output $z_i$ in $i$-th layer is the input $x_{i+1}$ for $\{i+1\}$-th layer. And the output of deep feature is the output of the last transformer layer as
\begin{equation}
\begin{aligned}
  Out_{deep} = z_l.
  \label{eq:outd}
\end{aligned}
\end{equation}

\subsection{Broad Attention}
\begin{figure}[t]
  \centering
   \includegraphics[width=0.9\linewidth]{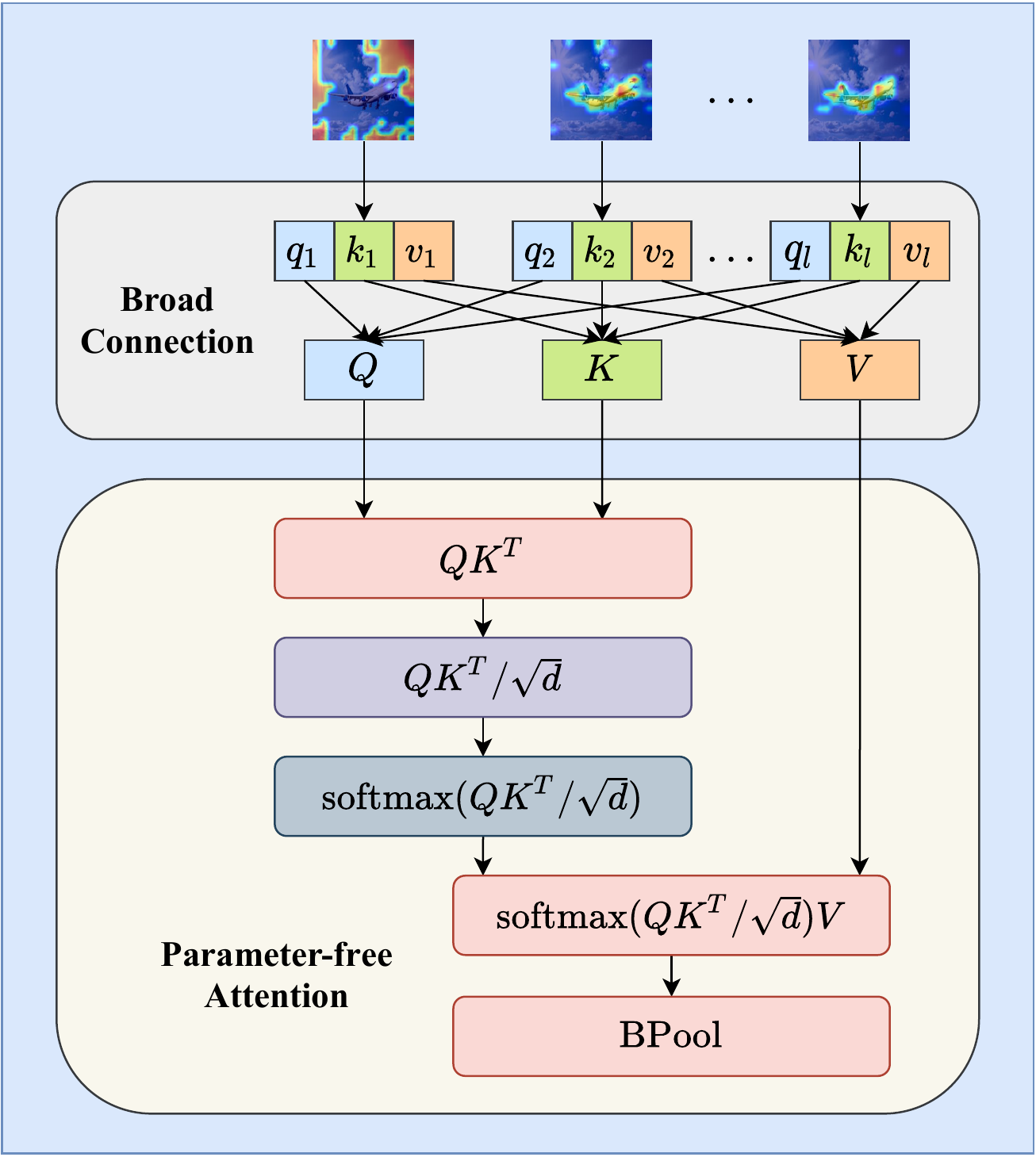}
   \caption{Illustration of broad attention block, including broad connection and parameter-free attention.}
   \label{fig:att_d}
\end{figure}

As the critical point of BViT architecture, broad attention block promotes the transmission
and integration of information from different layers using the broad connection, and focuses on helpful information hierarchically via parameter-free self-attention. Fig.~\ref{fig:att_d} introduces its detailed mechanism.

\textbf{Broad connection}: The broad connection promotes the transmission and integration of information flow by enhancing the path connection of attention, detailed as follows. For better understanding, we first give the equation of MHSA operation with $x_i \in \mathbb{R}^{ (N+1) \times D}$ as input
\begin{equation}
\begin{aligned}
  \text{MHSA}(x_i) & = \text{Atten}(\text{To\_qkv}(x_i))w^o \\
            & = \text{Atten}(q_i, k_i, v_i)w^o\\
            & = \text{softmax}(\frac {q_ik_i^T}{\sqrt{d_q}})v_iw^o,\\
    \text{and}~q_i & = [[q_i^1],[q_i^2], \dots, [q_i^h]], q_i^j \in \mathbb{R}^{(N+1)\times d_q}\\
    k_i & = [[k_i^1],[k_i^2], \dots, [k_i^h]], k_i^j \in \mathbb{R}^{(N+1)\times d_k} \\
    v_i& = [[v_i^1],[v_i^2], \dots, [v_i^h]], v_i^j \in \mathbb{R}^{(N+1)\times d_v}
  \label{eq:att}
\end{aligned}
\end{equation}
where $\rm To\_qkv$ contains linear projection, chunk and rearrange, $q_i^j$, $k_i^j$ and $v_i^j$ are the query, key and value of $j$-th head in $i$-th layer, $j \in [1,h]$. $\frac {1}{\sqrt{d_q}}$ is the scaling factor, 
and $w^o$ is the weight matrix of the second linear projection in MHSA block.
Then we respectively concatenate queries, keys and values of all MHSA blocks in different transformer layers as below
\begin{equation}
\begin{aligned}
  Q = [q_1, q_2,\dots , q_l], Q \in \mathbb{R}^{(N+1)\times h \times (l \times d_q)} \\
  K = [k_1, k_2,\dots , k_l], K \in \mathbb{R}^{(N+1)\times h \times (l \times d_k)} \\
  V = [v_1, v_2,\dots , v_l], V \in \mathbb{R}^{(N+1)\times h \times (l \times d_v)}
  \label{eq:qkv}
\end{aligned}
\end{equation}
where $q_i$, $k_i$, and $v_i$ are query, key, and value of $i$-th layer in Eq. (\ref{eq:att}). $Q$, $K$, and $V$ are concatenated queries, keys, and values accordingly. $l$ is the number of transformer layers. 

As can be seen in Fig.~\ref{fig:att_d}, the broad connection adequately integrates the features of each transformer layer. Intuitively, the connected $Q$, $K$, and $V$ express the attention information in different layers across the model, while $q_i$, $k_i$ and $v_i$ of each layer are only concerned with the attention information of a single layer. Since the increase of the path connection of attention in different layers, $Q$, $K$, and $V$ contain rich information and are more conducive to extracting helpful information.

In addition to the acquisition of information in different layers, another advantage of broad connection is the enhanced flow of information and gradients throughout the architecture, which facilitates its training. Each layer can access the gradient of the loss function directly, which helps to train a deeper network architecture.

\textbf{Parameter-free attention}: The parameter-free attention handles the integrated information via self-attention to jointly focus on helpful information and extract their relationship. Without introducing linear projection, we directly pay attention to available $Q$, $K$, and $V$. Thus broad attention does not bring extra learnable parameters, only slightly increasing the computational complexity. The implementation details of parameter-free attention $\text{Atten}_{\text{pf}}$ are as follows.

We perform self-attention on the concatenated queries $Q$, keys $K$, and values $V$, as we do in Eq. (\ref{eq:att}) except for linear projection. The specific formula is given below
\begin{equation}
\begin{aligned}
  \text{Atten}_{\text{pf}}(Q, K, V) & = \text{softmax}(\frac {QK^T}{\sqrt{d}})V \\
                   & = \text{softmax}(\frac {\sum_{i=1}^{l}{q_ik_i^T}}{\sqrt{d}})V \\
                   & = \text{softmax}(\frac {\sum_{i=1}^{l}{q_ik_i^T}}{\sqrt{d}})[v_1, v_2,\dots , v_l],
  \label{eq:dots}
\end{aligned}
\end{equation}
where $\frac {1}{\sqrt{d}}$ is the scaling factor, and $d$ is the hidden dimension of transform layer. From Eq.~(\ref{eq:dots}), we can conclude that parameter-free attention not only leverages the concatenated values (i.e. $[v_1, v_2,\dots , v_l]$) of the different layers, but also integrates attention weights of the different layers (i.e. $\sum_{i=1}^{l}{q_ik_i^T}$). 
The concatenation of attention information in different layers leads to the inconsistent dimension between $Out_{deep}$ and output of $\text{Atten}_{\text{pf}}(Q, K, V)$. We introduce a pooling operator $\rm BPool$ for broad feature $Out_{broad}$ to deal with the above problem. $\rm BPool$ performs a 1D adaptive average pooling on the inconsistent dimension of $\text{Atten}_{\text{pf}}(Q, K, V)$. The output feature of broad attention can be expressed as
\begin{equation}
\begin{aligned}  
  Out_{broad} = \text{BPool}(\text{Atten}_{\text{pf}}(Q, K, V), \{d_p\}),
  \label{eq:outde}
\end{aligned}
\end{equation}
where $d_p$ is the dimension of deep feature $Out_{deep}$.
Focusing on attention information in different layers across the model, parameter-free attention has a tremendous potential to extract the critical information and their structural relationship, which contributes to the performance of the model on vision tasks without additional model training burden.

With combination of deep feature output $Out_{deep}$ in Eq. (\ref{eq:outd}) and broad feature output $Out_{broad}$ in Eq. (\ref{eq:outde}), we can derive final output $Out$ is computed by 
\begin{equation}
\begin{aligned}
  Out = Out_{deep} + \gamma \ast Out_{broad},
  \label{eq:out}
\end{aligned}
\end{equation}
where $\gamma$ is the coefficient factor, which can be used to adjust the weight of two different types of feature.

In a nutshell, BViT backbone uses multiple transformer layers to process the image input into the more understandable feature for the model. Based on this, broad attention block enables the model to jointly attend to information from different representation spaces at different transformer layers. By extracting attention information in different layers rather than a single layer, broad attention can obtain wealthier information, which allows us to pay attention to vital information and ignore redundant information. In the broad attention block, the broad connection is responsible for enhancing the path connection of attention in different layers to facilitate the transmission
and integration of information. Parameter-free attention is responsible for attending to vital information from different layers and extracting their relationship. Consequently, benefiting from the introduction of broad attention block, BViT can concentrate better on the significant information which improves the classification accuracy of the model. Moreover, the inclusion of broad attention does not bring additional learnable parameters, which enables the convenient application on attention-based models.

%% file: experiment.tex
\section{Experiments}
\label{exp}
 We conduct the following experiments with BViT on ImageNet~\cite{deng2009imagenet} image classification and downstream tasks (i.e. CIFAR10/100~\cite{krizhevsky2009learning}). 
We first give the datasets introduction and experimental setup. Next, we perform ablation studies to validate the importance of elements of broad attention block, including coefficient factor and concatenated value $V$. Then, we compare the proposed BViT architecture with state-of-the-art works and apply broad attention to several remarkable ViT models, such as Swin Transformer~\cite{liu2021swin}, T2T-ViT~\cite{yuan2021tokens} and LVT~\cite{yang2022lite}, to verify the generality of broad attention. Finally, we analyze the visualization results of BViT in detail.

\subsection{Setup}
To validate the performance of our proposed model, we use the ImageNet dataset~\cite{deng2009imagenet}, which contains about 1.3M training data and 50K validation data with various 1000 object classes. Furthermore, we transfer pre-trained BViT on ImageNet to downstream
datasets, such as CIFAR10/100~\cite{krizhevsky2009learning}, which are small-scale image classification datasets with 50K training data and 10K test data.

\textbf{Model Variants}: To demonstrate the performance of BViT on the image classification task, we construct two models with different sizes, including BViT-5M and BViT-22M. The architecture specifications of BViT are shown as follows. The number of heads and dimensions of the transformer layer are 3 and 192 for BViT-5M, while 6 and 384 for BViT-22M.
Besides, several architectural parameters remain consistent across the two models. For instance, depth and MLP ratio are set to 12 and 4 respectively. BViT does not increase the trainable parameters compared to vanilla ViT without broad attention block. The increase in FLOPs due to broad attention block is tiny (i.e. $10^{-5}G$) and is therefore negligible. The coefficient factor $\gamma$ is simply set to 1 in the following experiments.

\textbf{Training setting}: Our training setting mostly follows DeiT~\cite{touvron2021training}. For all model variants, the input image resolution is 224 $\times$ 224. And we train models for 300 epochs, employing Adamw~\cite{kingma2014adam} optimizer and using cosine decay learning rate scheduler. Due to the limitation of computing resources, the batch size of BViT-5M and BViT-22M are 1280 and 512, respectively. Learning rate varies with batch size, same as DeiT~\cite{touvron2021training}. The weight decay of 0.05 is applied. The step of warmup is set to 5000. Further, We employ a majority of the augmentation and regularization strategies of DeiT~\cite{touvron2021training} in training, such as RandAugment~\cite{cubuk2020randaugment}, Mixup~\cite{zhang2017mixup}, Cutmix~\cite{yun2019cutmix}, Random Erasing~\cite{zhong2020random}, Stochastic Depth~\cite{huang2016deep} and Exponential Moving Average ~\cite{polyak1992acceleration}, except for Repeated Augmentation~\cite{hoffer2020augment} which cannot deliver significant performance boosts.

\textbf{Fine-tuning setting}: Our fine-tuning setting mostly follows ViT~\cite{DBLP:conf/iclr/DosovitskiyB0WZ21}. Using SGD optimizer with a momentum of 0.9, we pre-train our models at resolution 224 $\times$ 224. Then we fine-tune each model with batch size of 512. The training steps for CIFAR10/100 are 17000.

\subsection{Ablation Study}

\subsubsection{\textbf{Coefficient factor}}
The coefficient factor adjusts the weight of the broad and deep features as shown in Eq. (\ref{eq:out}). To discuss the effectiveness of different coefficient factors, we conduct the ablation study with different coefficient factors on ImageNet. Concretely, we choose 0.2, 0.4, 0.6, 0.8, and 1 as coefficient factor candidates.

\begin{figure}[t]
  \centering
   \includegraphics[width=1.0\linewidth]{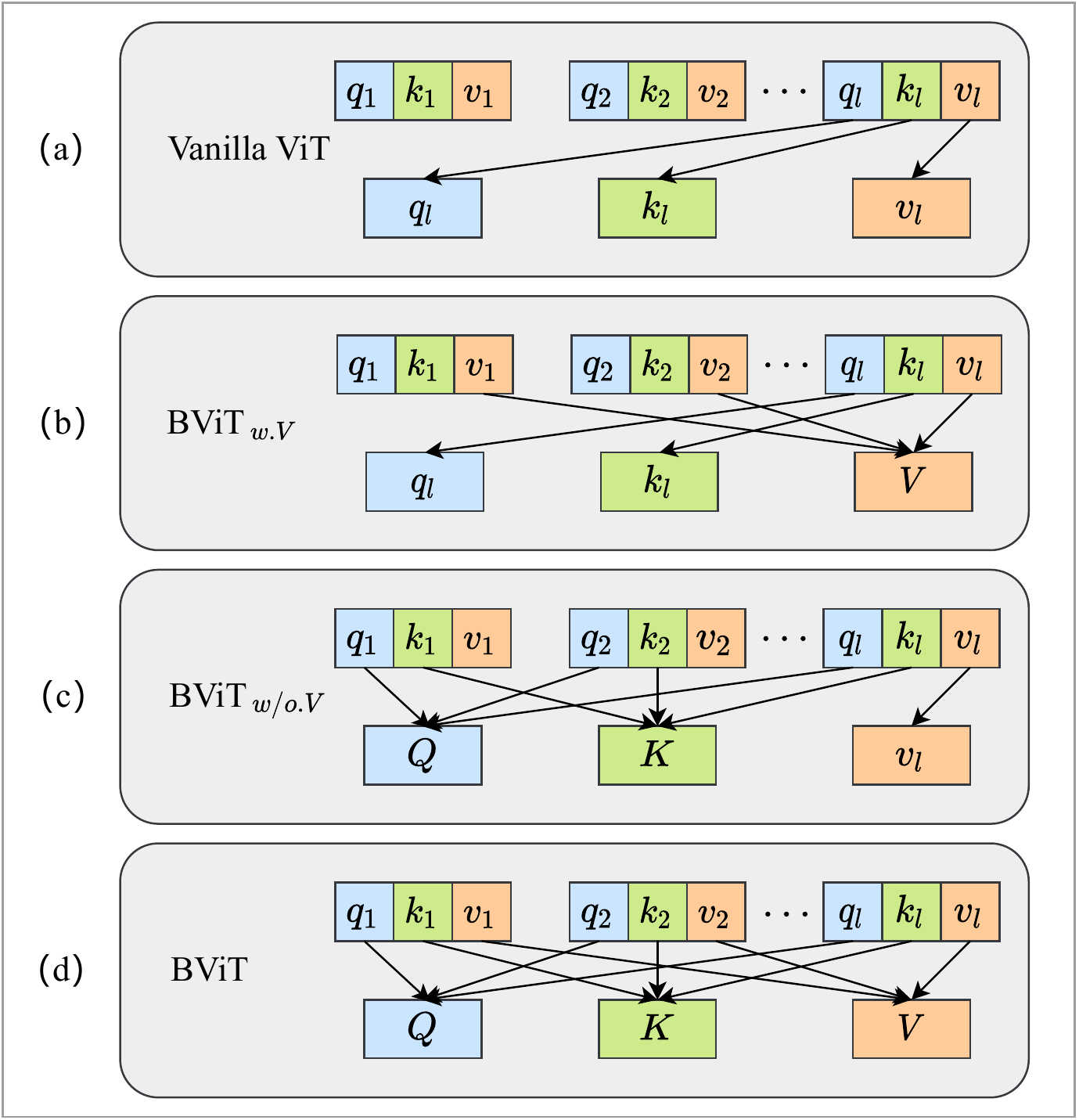}
   \caption{Architectural details of the four models in ablation study.}
   \label{fig:abl}
\end{figure}
As shown in Table~\ref{tab:ablac}, all coefficient factor candidates result in an improvement of over 2\%, which confirms the powerful effectiveness of broad attention. The greatest increase (i.e. 3.1\%) is achieved with the coefficient factor of 0.6. However, considering the slight variation in the results due to the randomness of the training, we deem it acceptable to choose any coefficient factor candidates. Thus we simply set the coefficient factor to 1 for the rest of the experiments. 
\begin{table}[h]
\renewcommand\arraystretch{1.2}
\centering
\caption{Ablations for different coefficient factors on ImageNet.}
\label{tab:ablac}
\begin{tabular*}{5.8cm}{lcc}
\hline
\textbf{Network} & \textbf{coefficient factor}  & \textbf{Top1 Acc}\\

\hline
DeiT-Ti~\cite{touvron2021training} & 0    & 72.2  \\
BViT$_{0.2}$                       & 0.2  & 74.3  \\
BViT$_{0.4}$                       & 0.4  & 74.8  \\
BViT$_{0.6}$                       & 0.6  & \textbf{75.3} \\
BViT$_{0.8}$                       & 0.8  & 74.5  \\
BViT$_1$                           & 1    & 75.0 \\
\hline
\end{tabular*}
\end{table}

\subsubsection{\textbf{With $V$ v.s. Without $V$}}
As mentioned above, the power of BViT stems from its broad attention block which jointly pays attention to the information in different layers. The helpful attention information in different layers consists of two components, i.e. concatenated values $V$ and aggregated attention weights $QK^T$. In order to discuss their respective effectiveness, we conduct an ablation study with four different architectures as shown in Fig.~\ref{fig:abl}. 
To discuss the different components of the information, the four architectures differ mainly in the integration of information in broad connection.
The four architectures are 1) Vanilla ViT as shown in Fig.~\ref{fig:abl} (a), without broad attention, it only utilizes the attention in the last transformer layer, i.e. DeiT\cite{touvron2021training}. 2) BViT$_{w.V}$ as shown in Fig.~\ref{fig:abl} (b), with concatenated values $V$, it performs broad attention using concatenated values $V$ in all transformer layers and attention weight $q_lk_l^T$ of last transformer layer;  3) BViT$_{w/o.V}$ as shown in Fig.~\ref{fig:abl} (c), without concatenated values $V$, it delivers broad attention via $v_l$ in last transformer layer and broadly connected attention weights $QK^T$; 4) BViT as shown in Fig.~\ref{fig:abl} (d), it focuses on both concatenated values $V$ and aggregated attention weights $QK^T$.

Ablation experiments on ImageNet~\cite{deng2009imagenet} are reported in Table~\ref{tab:abla}. As a comparison baseline without broad attention block, DeiT-Ti~\cite{touvron2021training} has the same architectural settings for the transformer layers with our BViT-5M. As shown in Table~\ref{tab:abla}, it is apparent that all three architectures with broad connection yield significant improvement. The heaviest improvement is in BViT which implements both concatenated values $V$ and aggregated attention weights $QK^T$. Besides, aggregated attention weights $QK^T$ deliver a slightly greater performance improvement than concatenated values $V$. Nevertheless, all three methods with broad connection can deliver an accuracy improvement of more than 2\%. Thus the choice of practical application can be carried out according to the architectural requirement.

\begin{table}[ht]
\renewcommand\arraystretch{1.2}
\centering
\caption{Ablations on components for broad attention on ImageNet. All three architectures with broad connection provide significant performance improvement.}
\label{tab:abla}
\begin{tabular*}{5.8cm}{lccc}
\hline
\multirow{2}*{\textbf{Network}} & \multicolumn{2}{c}{\textbf{Broad Attention}}  & \multirow{2}*{\textbf{Top1 Acc}}\\
\cline{2-3}
  & $QK^T$ & $V$ & \\
\hline
DeiT-Ti~\cite{touvron2021training} & $\times$       & $\times$ & 72.2  \\
BViT$_{w.V}$                               &  $\times$       & $\checkmark$  & 74.2 \\
BViT$_{w/o.V}$                             & $\checkmark$   & $\times$    & 74.4  \\
BViT                          &         $\checkmark$   & $\checkmark$  & \textbf{75.0}  \\
\hline
\end{tabular*}
\end{table}

\subsection{Image Classification on ImageNet}

\begin{table*}[ht]
\renewcommand\arraystretch{1.2}
\centering
\caption{Performance comparison with state-of-the-art models on ImageNet, including transformer-based models, CNN-based models, MLP-based models and hybrid models. We group models based on the size of their parameters and method type. The proposed BViT-5M outperforms all the manual-designed ViT methods without broad attention at less 2.6\% with about 5M parameters. In these models with about 22M parameters, the proposed BViT-22M has also outperformed the classical ViT method ViT-S \cite{DBLP:conf/iclr/DosovitskiyB0WZ21} about 3\%, even exceeds prevalent ViT method Swin-S\cite{liu2021swin}. BLVT and BSwin even deliver state-of-the-art performance among transformer-based models.}
\label{tab:imagenet}
\begin{tabular*}{13.6cm}{lcccccc}
\hline
\textbf{Network} & \textbf{Resolution} & \textbf{Params} & \textbf{FLOPs} & \textbf{Top1 Acc} & \textbf{Method Type} & \textbf{Design Type}\\
\hline  
MobileNetV3$_{Large0.75}$~\cite{howard2019searching} & $224^2$  & 4.0M    & 0.16G    & 73.3    & CNN           & Auto   \\
ResNet-18~\cite{he2016deep}                         & $224^2$  & 12.0M   & 1.8G     & 69.8    & CNN           & Manual \\
EfficietNet-B0~\cite{tan2019efficientnet}           & $224^2$  & 5.4M    & 0.39G    & 77.1    & CNN           & Auto   \\
\hline
gMLP-Ti~\cite{liu2021pay}                           & $224^2$  & 6.0M    & 1.4G     & 72.3    & MLP           & Manual \\
DeiT-Ti~\cite{touvron2021training}                  & $224^2$  & 5.7M    & 1.2G     & 72.2    & Transformer   & Manual \\
T2T-ViT-7~\cite{yuan2021tokens}                     & $224^2$  & 4.2M    & 0.9G     & 71.2    & Transformer   & Manual \\
% AutoFormer-Ti~\cite{chen2021autoformer}             & $224^2$  & 5.7M    & 1.3G     & 74.7    & Transformer   & Auto   \\
LVT~\cite{yang2022lite}                              & $224^2$  &5.5M   & 0.9G    & 74.8    & Transformer   & Manual \\
\textbf{BT2T-ViT-7(Ours)}                   & $224^2$  & 4.2M    & 0.9G     & 73.4    & Transformer   & Manual \\
\textbf{BViT-5M (Ours)}                              & $224^2$  & 5.7M    & 1.2G     & 75.0   & Transformer   & Manual \\
\textbf{BLVT(Ours)}        & $224^2$  & 5.5M   & 0.9G   & \textbf{76.0}   & Transformer   & Manual \\
\hline\hline
ResNet-50~\cite{he2016deep}                         & $224^2$  & 25.5M   & 4.1G     & 79.1    & CNN           & Manual \\
Inception v3~\cite{szegedy2016rethinking}           & $224^2$  & 27.2M   & 5.7G     & 77.4    & CNN           & Manual \\
RegNetY-4G~\cite{radosavovic2020designing}          & $224^2$  & 21.4M   & 4.0G     & 80.0    & CNN           & Auto   \\
EfficietNet-B4~\cite{tan2019efficientnet}           & $380^2$  & 19.3M   & 4.2G     & 82.9    & CNN           & Auto   \\
BoTNet-S1-59~\cite{srinivas2021bottleneck}          & $224^2$  & 33.5M   & 7.3G     & 81.7    & CNN + Trans   & Manual \\
\hline
gMLP-S~\cite{liu2021pay}                            & $224^2$  & 20.0M   & 4.5G     & 79.6    & MLP           & Manual \\
DeiT-S~\cite{touvron2021training}                   & $224^2$  & 22.1M   & 4.7G     & 79.9    & Transformer   & Manual \\
ViT-S/16~\cite{DBLP:conf/iclr/DosovitskiyB0WZ21}    & $384^2$  & 22.1M   & 4.7G     & 78.8    & Transformer   & Manual \\
Swin-T~\cite{liu2021swin}                           & $224^2$  & 29.0M   & 4.5G     & 81.3    & Transformer   & Manual \\
T2T-ViT-14~\cite{yuan2021tokens}                     & $224^2$  & 21.4M    & 4.8G     & 80.6    & Transformer   & Manual \\
% AutoFormer-S~\cite{chen2021autoformer}              & $224^2$  & 22.9M   & 5.1G     & 81.7    & Transformer   & Auto   \\
\textbf{BViT-22M (Ours)}                              & $224^2$  & 22.1M   & 4.7G     & 81.6    & Transformer   & Manual \\
\textbf{BSwin-T(Ours)}       & $224^2$  & 29.0M  & 4.5G    & 82.0  & Transformer   & Manual \\
\textbf{BSwin-S(Ours)}      & $224^2$  & 50.0M   & 8.7G    & \textbf{84.2}    & Transformer   & Manual \\
\hline
\end{tabular*}
\end{table*}
Table~\ref{tab:imagenet} presents the performance comparison to various types of architectures on ImageNet~\cite{deng2009imagenet}, including transformer-based models, CNN-based models, MLP-based models and hybrid models. The double line divides the experimental results into two blocks according to the number of parameters. The single line groups the models based on whether the method type contains CNN. In addition, the performance of other prevailing models with broad attention (i.e. BT2T-ViT, BLVT and BSwin) is included in Table~\ref{tab:imagenet}. In particular, BT2T-ViT, BLVT and BSwin are T2T-ViT, LVT and Swin with broad attention respectively.

Among manually designed transformer-based models, our BViT-5M achieves outstanding performance, outperforming VIT~\cite{DBLP:conf/iclr/DosovitskiyB0WZ21}, DeiT~\cite{touvron2021training} with approximately 2\% improvement, and even Swin Transformer~\cite{liu2021swin} which leads in various vision tasks. We also exceed gMLP~\cite{liu2021pay} by about 2\%, which delivers top results in the MLP-based models.
Further, BLVT and BSwin deliver state-of-the-art performance among transformer-based models.

Vision transformer is the newly promoted visual architecture, whose development is much less time-honored than CNN. The performance of transformer-based models in the visual field is still slightly inferior to that of CNN. Nevertheless, our BViT outperforms the partial CNN-based model, such as the classic and industry-known ResNet~\cite{he2016deep}, RegNetY~\cite{radosavovic2020designing}. Specifically, our BViT-22M surpasses ResNet-50 with a 2.5\% raise using fewer parameters, which considerably boosts the ImageNet classification task. 
% Further, we demonstrate the performance of attention-based models with broad attention in Table~\ref{tab:imagenet}, which has clear advantages.

To sum up, the robust performance of BViT proves that broad attention block is effective for capturing key features. By focusing on the attention information in different layers, BViT achieves innovation in the vision transformer. The innovation can contribute to other developed vision transformer architectures.

\subsection{Downstream tasks on CIFAR10/100}
\begin{table*}[ht]
\renewcommand\arraystretch{1.2}
\centering
\caption{Evaluation of transfer learning on downstream datasets. We transfer pre-trained BViT-22M on ImageNet to CIFAR10/100. BViT-22M takes $224\times224$ images during training and fine-tuning, and the accuracy of BViT exceeds ViT with higher resolution.}
\label{tab:down}
\begin{tabular*}{12.3cm}{lcccccc}
\hline
\textbf{Network} & \textbf{Resolution} & \textbf{Params} & \textbf{FLOPs} & \textbf{CIFAR10} & \textbf{CIFAR100} & \textbf{Method Type}\\
\hline
ViT-B/16~\cite{DBLP:conf/iclr/DosovitskiyB0WZ21}    & $384^2$  & 86.0M   & 18G      & 98.1    & 87.1   & Transformer \\
ViT-L/16~\cite{DBLP:conf/iclr/DosovitskiyB0WZ21}    & $384^2$  & 307.0M   & 190.7G      & 97.9    & 86.4   & Transformer \\
ResMLP-12~\cite{touvron2021resmlp}                          & $224^2$  & 15.0M   & 3.0G      & 98.1    & 87.0           & MLP \\
ResMLP-24~\cite{touvron2021resmlp}                          & $224^2$  & 30.0M   & 6.0G      & 98.7    & 89.5           & MLP \\
\hline
\textbf{BViT-22M (Ours)}                              & $224^2$  & 22.1M   & 4.7G     & \textbf{98.9}    & \textbf{89.9}   & Transformer \\
\hline
\end{tabular*}
\end{table*}
With the purpose of investigating the transferability of BViT, We evaluate the BViT-22M on downstream datasets via transfer learning. With BViT-22M being pre-trained on ImageNet~\cite{deng2009imagenet}, Table~\ref{tab:down} exhibits the comparison results on CIFAR10/100 that includes our BViT and other brilliant networks such as ViT~\cite{DBLP:conf/iclr/DosovitskiyB0WZ21} and ResMLP~\cite{touvron2021resmlp}. ResMLP is selected as the comparison model because it is fine-tuned at the resolution of $224\times224$, while most models choose $384\times384$.

From the results in Table~\ref{tab:down}, it is clear that compared to both transformer-based and MLP-based models, BViT achieves better classification accuracy on CIFAR10/100 with fewer parameters.
In general, our novel BViT maintains sound performance on downstream tasks, thus confirming its strength in the field of computer vision.

\subsection{Generalization Study}
The novel broad attention is generic due to its structural design, which can be implemented to improve the performance of attention-based models. In order to verify the generalization of broad attention, we introduce broad attention to three excellent models, such as T2T-ViT~\cite{yuan2021tokens}, LVT~\cite{yang2022lite} and Swin Transformer~\cite{liu2021swin}, and derive the BT2T-ViT, BLVT and BSwin.

The specific experimental settings are consistent among all models. We apply broad attention to attention-based models and fine-tune the model with pre-trained weights.  In fine-tuning, we train the model for 30 epochs with a batch size of 1024, a constant learning rate of $10^{-5}$ and a weight decay of $10^{-8}$. In the following, we will give experimental results.

\begin{table}[ht]
\renewcommand\arraystretch{1.2}
\centering
\caption{Performance comparison between excellent attention-based models and models with broad attention on ImageNet. Benefit from broad attention, all models show significant performance gains without extra parameters.}
\label{tab:ablage}
\begin{tabular*}{7.5cm}{lccc}
\hline
~&\textbf{Model}  &\textbf{Params} & \textbf{Top1 Acc}\\
\hline
\multirow{2}{*}{T2T-ViT~\cite{yuan2021tokens}} &T2T-ViT-7 &  4.2M & 71.7 \\
~ & \cellcolor{lightgray}BT2T-ViT-7& \cellcolor{lightgray} 4.2M & \cellcolor{lightgray}\textbf{73.4}$_{(\textcolor{red}{+1.7})}$\\
\hline
\multirow{2}*{LVT~\cite{yang2022lite}} &LVT& 5.5M & 74.8 \\
~ & \cellcolor{lightgray}BLVT&  \cellcolor{lightgray}5.5M & \cellcolor{lightgray}\textbf{76.0}$_{(\textcolor{red}{+1.2})}$\\
\hline
\multirow{4}*{Swin~\cite{liu2021swin}} &Swin-T& 29M & 81.3 \\
~ &  \cellcolor{lightgray}BSwin-T&  \cellcolor{lightgray}29M & \cellcolor{lightgray}\textbf{82.0}$_{(\textcolor{red}{+0.7})}$\\
~ & Swin-S&  50M & 83.0\\
~ & \cellcolor{lightgray}BSwin-S&  \cellcolor{lightgray}50M & \cellcolor{lightgray}\textbf{84.2}$_{(\textcolor{red}{+1.2})}$\\
\hline
\end{tabular*}
\end{table}

\subsubsection{\textbf{BT2T-ViT}}

The overview of T2T-ViT consists of Tokens-to-Token module and T2T-ViT backbone. We apply broad attention to T2T-ViT backbone. Since the dimension of each T2T Transformer layer in T2T-ViT backbone is consistent, the specific implementation of broad attention in BT2T-ViT is exactly identical to BViT.

Table~\ref{tab:ablage} presents the performance comparison between T2T-ViT-7~\cite{yuan2021tokens} and BT2T-ViT-7 on ImageNet~\cite{deng2009imagenet}. Benefiting from the exploitation of attention in different layers brought by broad attention, BT2T-ViT-7 exceeds original T2T-ViT-7 1.7\% without extra parameters.

\subsubsection{\textbf{BLVT}}

LVT~\cite{yang2022lite} consists of two novel self-attention blocks: Convolutional Self-Attention (CSA) and Recursive Atrous Self-Attention (RASA). 
We apply broad attention to RASA. Besides, considering the inconsistent dimension of the transformer block, i) we only connect the last block of each stage instead of all LVT Transformer blocks. and ii) we apply maximum pooling and reshaping to obtain the consistent dimension of RASA in different stages.

Table~\ref{tab:ablage} exhibits the comparison results on ImageNet~\cite{deng2009imagenet} that includes LVT~\cite{han2021transformer} and BLVT. Experimental results show that broad attention brings a 1.2\% performance improvement compared to the original LVT.

\subsubsection{\textbf{BSwin}}

Swin Transformer~\cite{liu2021swin} is a hierarchical Transformer, which models features with different scales in different stages. We broadly connect the outputs of shifted window-based self-attention in the last block of each stage to extract attention information in different layers. Similar to BLVT, there are also two differences from BViT, which are induced by the different scales of features in the hierarchical Swin Transformer. 
% The architectural details of the broad attention block are the same as BViT.

Table~\ref{tab:ablage} exhibits the comparison results on ImageNet~\cite{deng2009imagenet} that includes Swin-T/S~\cite{liu2021swin} and BSwin-T/S. It is clear that compared to the original Swin-T/S, BSwin-T/S achieves better classification accuracy on ImageNet without extra parameters.

\subsubsection{\textbf{Summary}}
To sum up, the above experimental results fully demonstrate that our broad attention is generic and can flexibly improve the performance of attention-based models. In a word, broad attention can be introduced to attention-based models as a generic mechanism. Furthermore, the outstanding performance of BT2T-ViT, BLVT and BSwin also proves that broad attention is effective for leveraging significant features. Paying attention to information from multiple layers, the broad attention helps to understand image representation for classification.

\subsection{ Visualization}
To further investigate the impact of broad attention on representation, we conduct three visualization experiments, including centered kernel alignment similarity, mean attention distance and attention map. CKA similarity is to discuss the influence of broad attention on the similarity of features in different layers. Mean attention distance can analyze the size of the attention area for BViT. The attention map demonstrates the representation of BViT from the output token to the input image.

\begin{figure}[!t]
\centering
\subfloat[]{\includegraphics[width=1.8in]{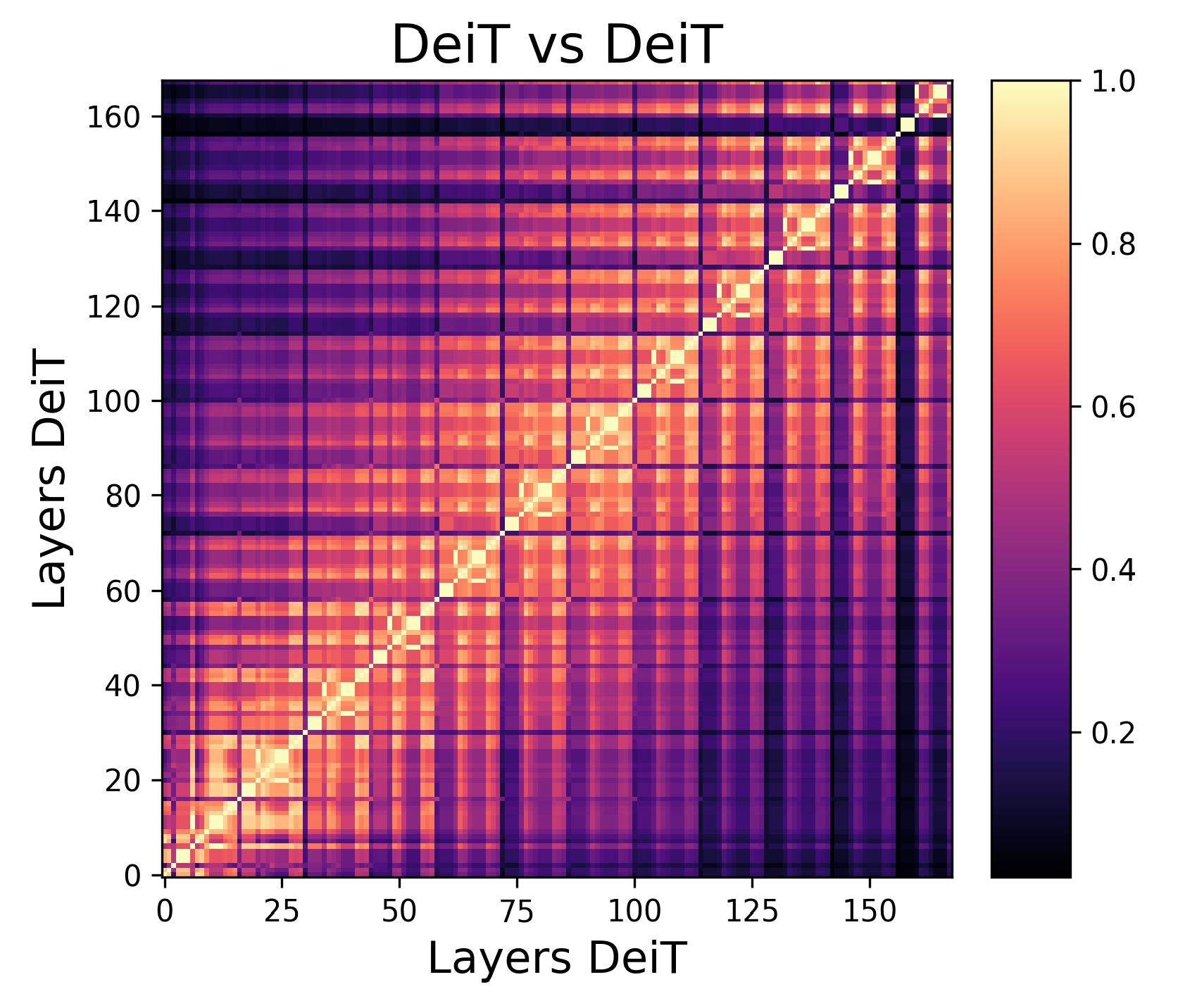}
\label{ckad}}
\subfloat[]{\includegraphics[width=1.8in]{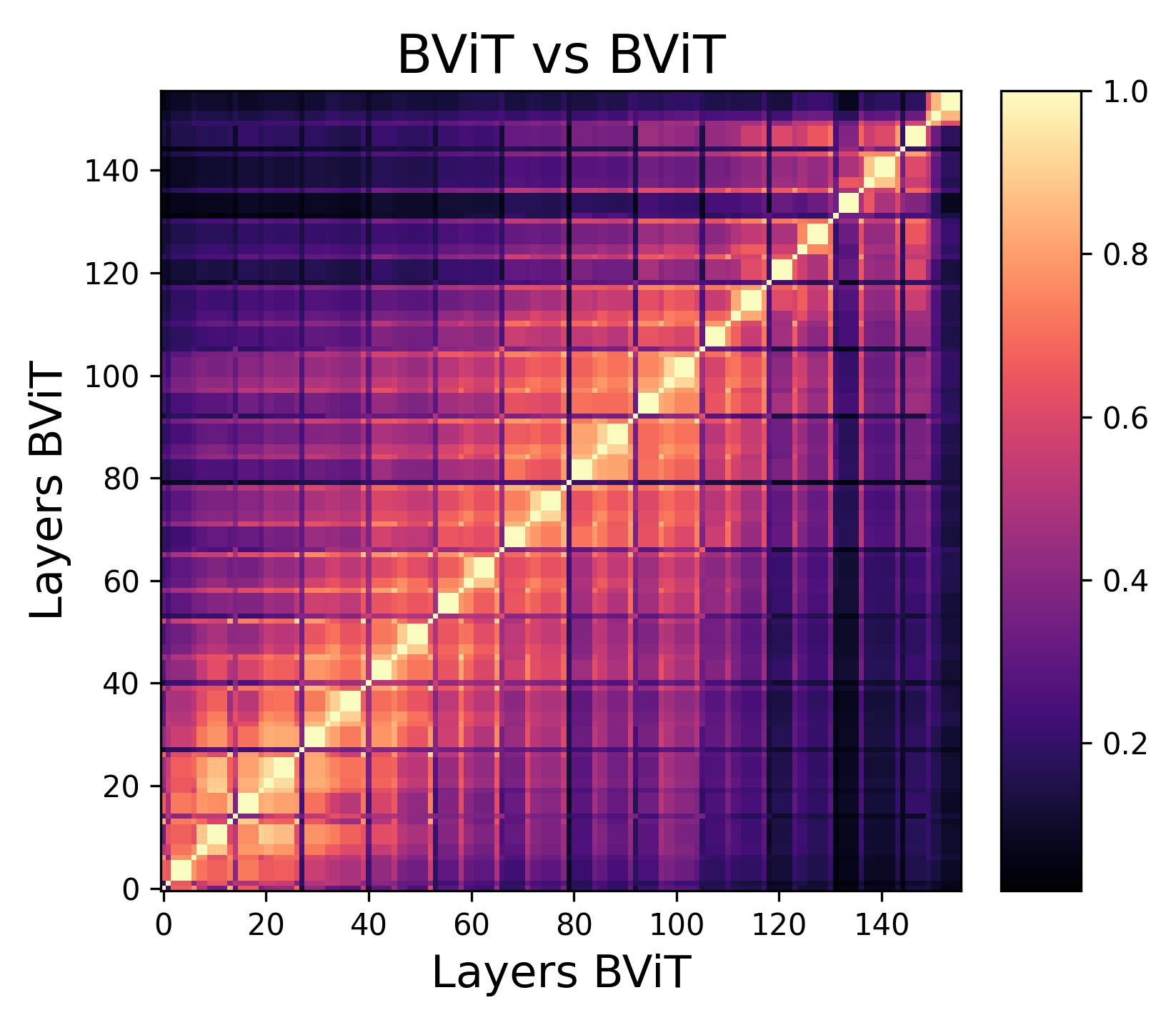}
\label{ckab}}
\caption{Representation similarity comparison via CKA: (a) DeiT~\cite{touvron2021training} (i.e. architecture without broad attention), (b) BViT. We present CKA similarities between all pairs of transformer layers for our BViT and DeiT. Both horizontal and vertical coordinates indicate the number of architectural layers, and the color indicates the scores of similarity. The lighter the color, the higher the scores of similarity. We randomly sample 1000 images from ImageNet~\cite{deng2009imagenet} dataset to compute CKA similarity scores. The heatmaps illustrate that BViT has smaller similarity scores between shallower and deeper layers than DeiT without broad attention.}
\label{cka}
\end{figure}
\subsubsection{\textbf{Centered Kernel Alignment Similarity}}
CKA~\cite{raghu2021vision} enables a quantitative measure of representation similarity within models. Researchers often employ it to explore the differences in representation learned by visual models, especially when studying the difference between ViT and CNN~\cite{raghu2021vision}. Moreover,~\cite{nguyen2020wide} states that similar features may mean redundancy in the model. Concretely, given the representation $X$ and $Y$ of two layers as inputs, we can derive Gram matrices $K=XX^T$ and $L=YY^T$. Then CKA can be computed as
\begin{equation}
\begin{aligned}  
  \text{CKA}(K,L) = \frac{\text{HSIC}(K,L)}{\sqrt{\text{HSIC}(K,K)\text{HSIC}(L,L)}},
  \label{eq:outpb}
\end{aligned}
\end{equation}
where HSIC is the Hilbert-Schmidt Independence Criterion~\cite{gretton2007kernel}. The representation similarity comparison between BViT and DeiT~\cite{touvron2021training} (i.e. architecture without broad attention) is shown in Fig.~\ref{cka}. It can be seen that the CKA similarity scores are smaller between shallower and deeper layers in BViT (Fig.~\ref{cka} (b)) than in DeiT, which means less model redundancy. Therefore, the design of broad attention helps extract and utilize features effectively, leading to better performance.
\begin{figure}[!t]
\centering
\subfloat[]{\includegraphics[width=1.6in]{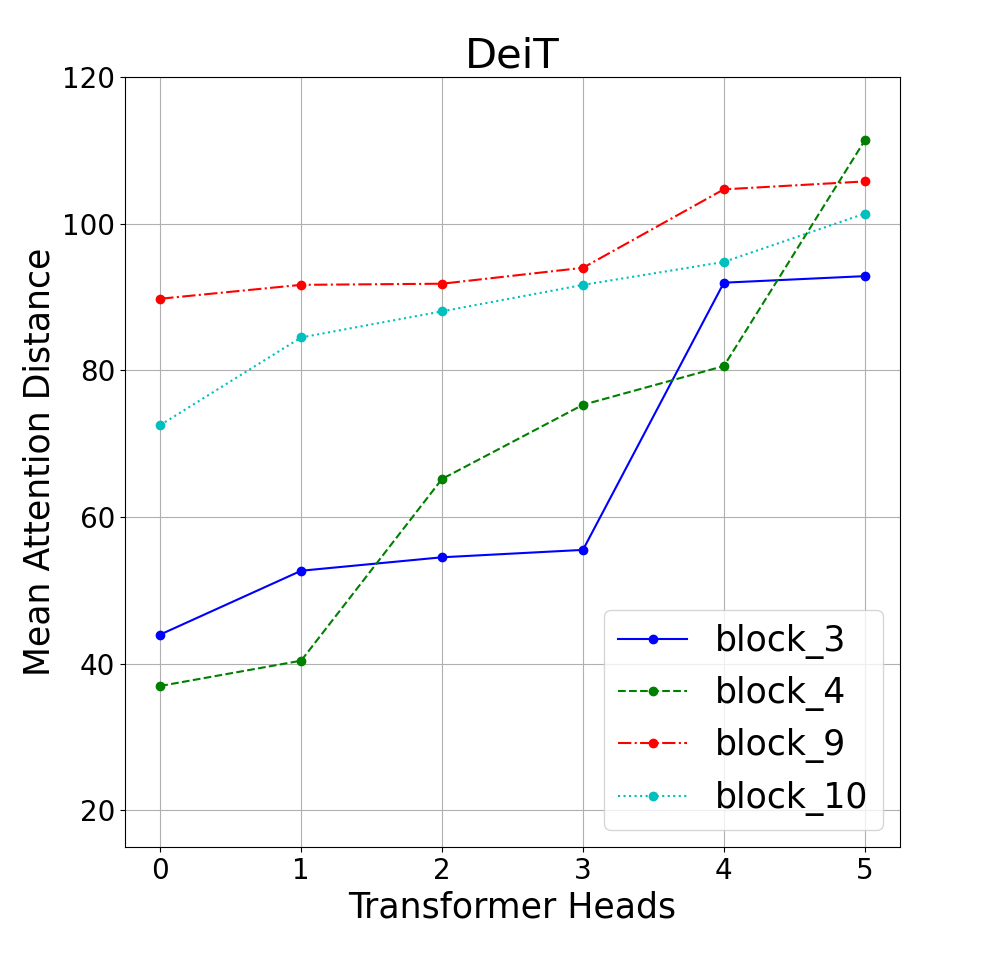}
\label{deitatt}}
\subfloat[]{\includegraphics[width=1.6in]{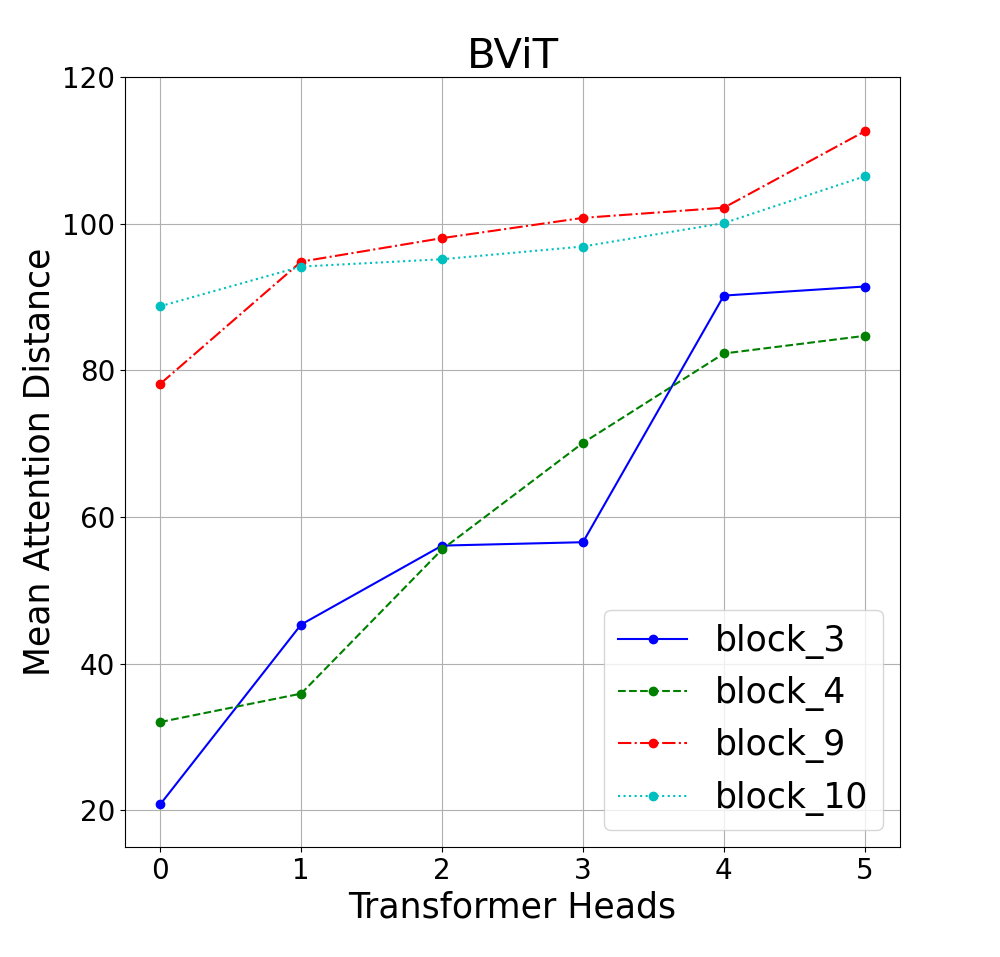}
\label{bvitatt}}
\caption{Mean attention head distance by 6 attention heads: (a) DeiT~\cite{touvron2021training} (i.e. architecture without broad attention), (b) BViT. The horizontal coordinate indicates the heads of attention and the vertical coordinate indicates the mean attention distance. Different lines indicate attention blocks of different layers. Following ViT~\cite{DBLP:conf/iclr/DosovitskiyB0WZ21}, we randomly sample 128 images from ImageNet~\cite{deng2009imagenet} dataset and calculate the mean distance between pixels with attention weights.}
\label{attdis}
\end{figure}
\subsubsection{\textbf{Mean Attention Distance}}
Mean Attention Distance is first proposed in ViT~\cite{DBLP:conf/iclr/DosovitskiyB0WZ21}. The computation of mean attention distance requires averaging the distance between the query pixel and other pixels, that is weighted by the attention weight. The research about the difference between ViT and CNN~\cite{raghu2021vision} demonstrates experimentally that the model with more local attention performs better on limited datasets.
As shown in Fig~\ref{attdis}, in order to figure out what influence broad attention brings to attention distance, we plot the mean attention head distance of the third, fourth, ninth and tenth blocks by sorted heads. The results show that our BViT (Fig.~\ref{attdis} (b)) attends more local information at the shallow layers. For example, in the third block's first head, the mean attention distance of DeiT is about 40 while BViT is about 20 (blue line). As stated in previous research~\cite{raghu2021vision}, more attention to local features may facilitate the learning of the model on the limited datasets. Thus our model achieves better classification accuracy on ImageNet~\cite{deng2009imagenet}.

\subsubsection{\textbf{Attention Maps}}
To illustrate the significance of broad attention block, we use Attention Rollout~\cite{abnar2020quantifying} to compute attention maps of transformer layers. 
Attention Rollout averages the attention weights of the model across all heads and then recursively multiplies the weight matrices of different transformer layers. 

\begin{figure*}[t]
  \centering
   \includegraphics[width=1.0\linewidth]{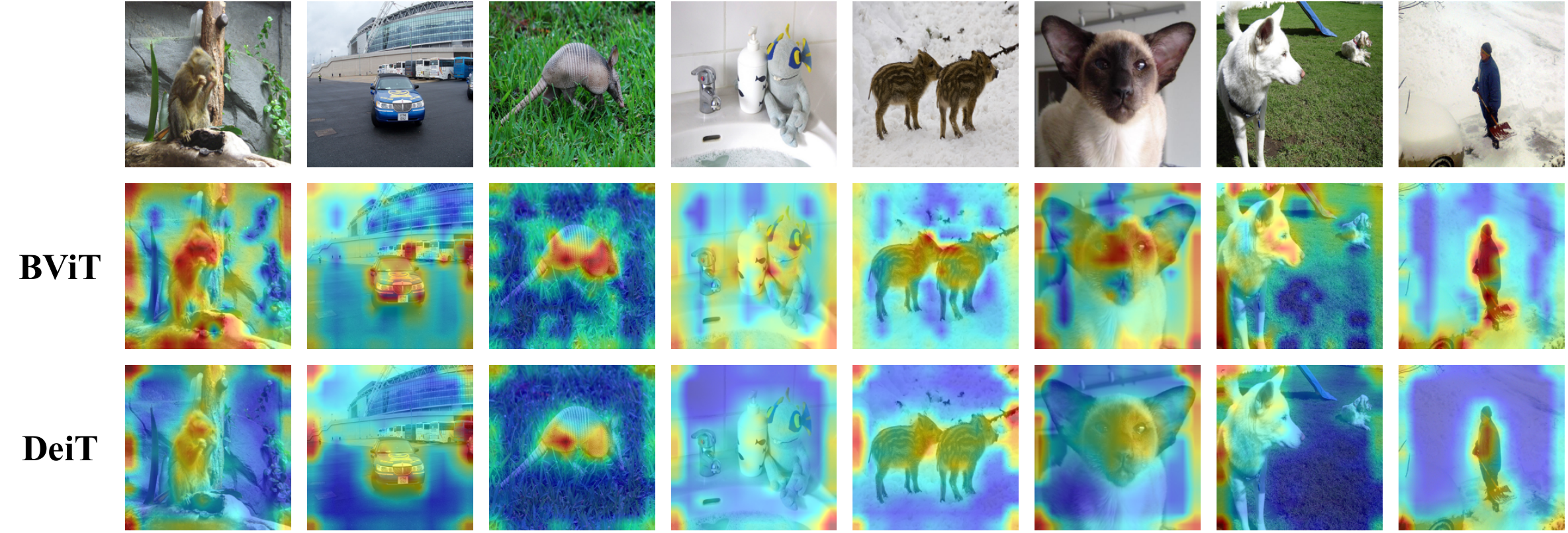}
   \caption{Comparison of attention maps between DeiT (i.e. architecture without broad attention) and our BViT. The attention maps of BViT focus more on the object to be classified than DeiT and that is positive for image recognition.}
   \label{fig:vua}
\end{figure*}

As shown in Fig.~\ref{fig:vua}, we visualize the attention maps of BViT and DeiT~\cite{touvron2021training} (i.e. architecture without broad attention). 
Visualization results show that utilization of the attention in different layers facilitates the spotting of the critical object. 
The attention maps of BViT pay more attention to the object to be recognized than DeiT.
The phenomenon provides an intuitive argument for the design of our broad attention mechanism, which reasonably improves the understanding of images by enhancing the exploitation of features. 

In summary, benefiting from the design of broad attention, our BViT can i) achieve effective features by preventing model redundancy, ii) deliver excellent performance on limited datasets (e.g. ImageNet) for more local attention, and iii) achieve better image understanding.

%% file: conclusion.tex
\section{Conclusion}
This paper proposes the Broad attention based Vision Transformer, called BViT. As the key element of BViT, broad attention consists of broad connection and parameter-free attention. Broad connection integrates attention information in different layers. Then parameter-free attention extracts effective features from the above integrated information and constructs their relationships. Furthermore, due to the novel broad attention block being directed at the existing attention, the proposed broad attention is generic to improve the performance of attention-based models.
Consequently, BViT achieves leading performance on vision tasks benefiting from rich and valuable information. On ImageNet, BViT arrives at state-of-the-art performance among transformer-based models with about 3\% boost to groundbreaking ViT. Then we transfer BViT-22M to downstream tasks (CIFAR10/100) that prove the robust transferability of the model.
Moreover, the implementation of broad attention on T2T-ViT, LVT and Swin Transformer also improves accuracy by more than 1\%, confirming the flexibility and effectiveness of our method.

As a key component of BViT, the broad attention block significantly improves the performance of the vision transformer on image classification. We expect to inspect its employment in natural language processing tasks. Further, we will explore the impact of different connection combinations of transformer layer outputs on performance via neural architecture search algorithm.